\documentclass{article}

\usepackage[final,nonatbib]{neurips_2023}

\usepackage[utf8]{inputenc} 
\usepackage[T1]{fontenc}    
\usepackage{hyperref}       
\usepackage{url}            
\usepackage{booktabs}       
\usepackage{amsfonts,amsmath}       
\usepackage{nicefrac}       
\usepackage{microtype}      
\usepackage{xcolor}         
\usepackage{lmodern}
\usepackage{graphicx}
\usepackage{comment}
\usepackage{caption}
\usepackage{subcaption}
\usepackage{wrapfig}

\newtheorem{definition}{Definition}

\usepackage{multirow}

\title{Assessment of the Reliablity of a Model's Decision by Generalizing Attribution to the Wavelet Domain}

\author{%
Gabriel Kasmi$^{1,2}$ \quad Laurent Dubus$^{2,3}$ \quad Yves-Marie Saint Drenan$^1$\quad
Philippe Blanc$^{1}$\\
$^1$MINES Paris, Université PSL Centre Observation Impacts Energie (O.I.E.)\\
$^2$RTE France \\
$^3$WEMC (World Energy \& Meteorology Council, UK) \\
$^1$\texttt{\{firstname.lastname\}@minesparis.psl.eu}\\
$^2$\texttt{\{firstname.lastname\}@rte-france.com}\\
}

\begin{document}

\maketitle

\begin{abstract}
Neural networks have shown remarkable performance in computer vision, but their deployment in numerous scientific and technical fields is challenging due to their black-box nature. Scientists and practitioners need to evaluate the {\it reliability} of a decision, i.e., to know simultaneously if a model relies on the {\it relevant} features and whether these features are {\it robust} to image corruptions. Existing attribution methods aim to provide human-understandable explanations by highlighting important regions in the image domain, but fail to fully characterize a decision process's reliability. To bridge this gap, we introduce the {\bf W}avelet s{\bf C}ale {\bf A}ttribution {\bf M}ethod (WCAM), a generalization of attribution from the pixel domain to the space-scale domain using wavelet transforms. Attribution in the wavelet domain reveals where {\it and} on what scales the model focuses, thus enabling us to assess whether a decision is reliable. Our code is accessible here: \url{https://github.com/gabrielkasmi/spectral-attribution}.
\end{abstract}

\section{Introduction}

Deep neural networks have become the standard for numerous computer vision applications. However, there is a growing consensus that these models cannot be safely deployed in real-world applications \cite{delseny_white_2021} as models are not reliable enough, owing to two reasons \cite{saria_tutorial_2019}. Firstly, the black-box nature of deep neural networks motivates using explainable AI (XAI) techniques to generate a human-understandable explanation of a model's decision \cite{hagras_toward_2018}. Secondly, distribution shifts \cite{koh_wilds_2021} are ubiquitous in real-world cases and cause models to fail unpredictibly \cite{hendrycks_benchmarking_2019}. Safely deploying deep learning in real-world settings requires at least tools that enable auditing the {\it relevance} and {\it robustness} to distribution shifts of a model's decision. 

Attribution methods \cite{carvalho_machine_2019}, which consist of identifying the most important features in the input, have improved the understanding of the decision process of deep learning models. On the other hand, Fourier analysis has been extensively used to analyze the robustness of models \cite{yin_fourier_2020}. \cite{wang_high-frequency_2020,chen_rethinking_2022,zhang_range_2022} showed that robust models rely on low-frequency components. Existing attribution methods only represent the model decision in the pixel (space) domain, while Fourier only provides a decomposition in the frequency (scale) domain. To the best of our knowledge, no work has yet expanded attribution in the space-scale domain, which enables the audit of both the relevance {\it and} robustness of a model's decision.

We introduce the {\bf W}avelet s{\bf C}ale {\bf A}ttribution {\bf M}ethod (WCAM). This novel attribution method represents a model decision in the space-scale (or wavelet) domain. The decomposition in the space-scale domain highlights which structural components (textures, edges, shapes) are important for the prediction, allowing us to assess the {\it relevance} of a decision process. Moreover, as scales correspond to frequencies, we simultaneously evaluate whether this decision process is robust. We discuss the potential of the WCAM for application in expert settings (e.g., medical imaging or remote sensing), show that we can quantify the robustness of a prediction with the WCAM and highlight concerning phenomenon regarding the consistency of the decision process of deep learning models.

\section{Related works}

\paragraph{Explainability} Explainability in computer vision typically quantifies the contribution of an image's pixel or region to a model's prediction. Saliency \cite{simonyan_deep_2014} was among the first methods to identify such regions. The approach used the model's gradients and the classification score. A line of works improved this approach: instead of using the model's gradients, other works used the model's activations to generate explanations. It is the principle behind the class activation map (CAM, \cite{zhang_understanding_2016}), which has also been further refined \cite{shrikumar_not_2017,selvaraju_grad-cam_2020,sundararajan_axiomatic_2017}. These methods quickly compute an explanation but require access to the model's gradients or activation. We often refer to these methods as "white-box" explanation methods. By contrast, "black-box" methods are model agnostic. Explanations are computed by perturbing (e.g., occluding parts of the image) the inputs and computing a score that reflects the model's sensitivity to the perturbation. The various proposed methods, e.g., Occlusion \cite{zeiler_visualizing_2013}, LIME \cite{ribeiro_why_2016}, RISE,  \cite{petsiuk_rise_2018}, Sobol \cite{fel_look_2021}, HSIC \cite{novello_making_2022} or EVA \cite{fel_dont_2023} differ in that they use different sampling strategies to explore the space of perturbations and can be seen as special cases of a more general approach based on Shapley values \cite{lundberg_unified_2017}. However, the main limitation of these methods is that they only explain {\it where} the model focuses and are therefore not informative enough in many settings where one wants to assess {\it what} the model sees \cite{achtibat_where_2022}. 

To begin addressing the {\it what}, \cite{fel_craft_2023} recently introduced CRAFT. This method combines matrix factorization for concept identification and Grad-CAM \cite{selvaraju_grad-cam_2020} for concept localization on the input image. Another line of work focused on identifying the most significant points in the training dataset through influence functions \cite{koh_understanding_2020}. However, such approaches require access to the model and the training data and are, therefore, hard to implement in applied settings. 

\paragraph{Frequency-centric perspective on model robustness} A line of works aimed at explaining the behavior of neural networks through the lenses of frequency analysis. Several works showed that convolutional neural networks (CNNs) are biased towards high frequencies \cite{wang_high-frequency_2020,yin_fourier_2020} and that robust methods tend to limit this bias \cite{zhang_range_2022,chen_rethinking_2022}. Other works highlighted a so-called spectral bias \cite{rahaman_spectral_2019,xu_frequency_2020,jo_measuring_2017}, showing that CNNs learn the input image frequencies from the lowest to the highest. More recently, \cite{wang_what_2023} leveraged Fourier analysis to characterize learning shortcuts \cite{geirhos_shortcut_2020}: this work showed that learning shortcuts are context-dependent as models tend to favor the most distinctive frequency to make a prediction.

These works showed that the decomposition of a model's decision in the Fourier (frequency) domain characterizes what models see on the input. However, the Fourier domain only represents the frequencies, while the Wavelet (space-scale) domain provides an assessment in both space and frequency simultaneously \cite{mallat_wavelet_1999}. Expanding attribution into the wavelet domain could provide a more comprehensive assessment of what models see on the input, enabling the practitioner to gauge whether the model's decision is reliable.

\section{Methods}

\subsection{Background}

\paragraph{Wavelet transform} A wavelet is an integrable function $\psi\in L^2 (\mathbb{R})$ with zero average, normalized and centered around 0. Unlike a sinewave, a wavelet is localized in space and in the Fourier domain. It implies that dilatations of this wavelet enable to scrutinize different frequencies (scales) while translations enable to scrutinize spatial location. To compute an image's (continuous) wavelet transform (CWT), one first defines a filter bank $ \mathcal{D}$ from the original wavelet $\psi$ with the scale factor $s$ and the 2D translation in space $u$. We have
\begin{equation}
        \mathcal{D} = \left\{
        \psi_{s,u}(x) = \frac{1}{\sqrt{s}}\psi\left(\frac{x-u}{s}\right)
    \right\}_{u\in\mathbb{R}^2,\;s\ge 0},
\end{equation}
where $\vert \mathcal{D}\vert =J$, and $J$ denotes the number of levels. The computation of the wavelet transform of a function $f\in L^2(\mathbb{R})$ at location $x$ and scale $s$ is given by
\begin{equation}\label{eq:wt}
    \mathcal{W}(f)(x,s) = \int_{-\infty}^{+\infty}
    f(u) \frac{1}{\sqrt{s}} \psi^*\left(\frac{x-u}{s}\right)
    \mathrm{d}u,
\end{equation}
which can be rewritten as a convolution \cite{mallat_wavelet_1999}. Computing the multilevel decomposition of $f$ requires applying \autoref{eq:wt} $J$ times with all dilated and translated wavelets of $\mathcal{D}$. \cite{mallat_theory_1989} showed that one could implement the multilevel dyadic decomposition of the discrete wavelet transform (DWT) by applying a high-pass filter $H$ to the original signal $f$ and subsampling by a factor of two to obtain the {\it detail} coefficients and applying a low-pass filter $G$ and subsampling by a factor of two to obtain the {\it approximation} coefficients. Iterating on the approximation coefficients yields a multilevel transform where the $j^{th}$ level extracts information at resolutions between $2^j$ and $2^{j-1}$ pixels. The detail coefficients can be decomposed into horizontal, vertical, and diagonal components when dealing with 2D signals (e.g., images).

\paragraph{Sobol sensitivity analysis} The Sobol sensitivity analysis consists of decomposing the variance of the output of a model into fractions that can be attributed to a set of inputs. Let $(X_1,\dots,X_K)$ be independent random variables and $\mathcal{K}= \{1,\dots, K\}$ denote the set of indices. Let $f$ be a model, $X$ an input, and $f(X)$ the model's decision (e.g., the output probability). We denote $f_\kappa = f_\kappa(X_\kappa)$ the partial contributions of the variables $(X_k)_{k\in\kappa}$ to the score $f(X)$. The Sobol-Hoeffding decomposition \cite{hoeffding_class_1992} decomposes the decision score $f(X)$ into summands of increasing dimension
\begin{equation}\label{eq:sobol-hoeffding}
    f(X) = f_\emptyset + \sum_{\kappa\in\mathcal{P}\left(\mathcal{K}\right)\backslash\{\emptyset\}} f_\kappa(X_\kappa),
\end{equation}
Where $f_\emptyset$ denotes the prediction with no features, $\mathcal{P}\left(\mathcal{K}\right)$ denotes the power set of $\mathcal{K}$ and $\emptyset$ the empty set. Then,  $\forall(u,v)\in\mathcal{K}^2$ such that $u\neq v$, $\mathbb{E}\left[f_u(X_u)f_v(X_v)\right] = 0$, we derive from \autoref{eq:sobol-hoeffding} the variance of the model's score
\begin{equation}\label{eq:variance}
    Var(f(X)) = \sum_{\kappa\in\mathcal{P}\left(\mathcal{K}\right)}Var(f_\kappa(X_\kappa)),
\end{equation}
\autoref{eq:variance} enables us to describe the influence of a subset $X_\kappa$ of features as the ratio between its own and the total variance. This corresponds to the first order {\bf Sobol index} given by 
\begin{equation}\label{eq:sobol-def}
S_\kappa = \frac{Var(f_\kappa(X_\kappa))}{Var(f(X))}.    
\end{equation}
$S_\kappa$ measures the proportion of the output variance $Var(f(X))$ explained by the subset of variables $X_\kappa$ \cite{sobol_sensitivity_1990}. Focusing on single features, $S_k$ captures the {\it direct} contribution of the feature $X_k$ to the model's decision. To capture the indirect effect, due to the effect of $X_k$ on the other variables, {\bf total Sobol indices} $S_{T_k}$ \cite{homma_importance_1996} can be computed as 
\begin{equation}
S_{T_k} = \sum_{\kappa\in\mathcal{P}\left(\mathcal{K}\right),\,k\in\kappa}S_\kappa.
\end{equation}
Total Sobol indices (TSIs) measure the contribution of the $k^{th}$ feature, taking into account both its {\it direct} effect and its {\it indirect} effect through its interactions with the other features. 

\paragraph{Efficient estimation of Sobol indices} As seen from \autoref{eq:sobol-def}, estimating the impact of a feature $k$ on the model's decision requires recording the partial contribution $f_k(X_k)$. This partial contribution corresponds to a {\it forward}. Estimating Sobol indices requires computing variances by drawing at least $N$ samples and computing $N$ forwards to estimate a first-order Sobol index $S_k$ of a single feature $k$. As we are interested in the TSI of a feature $k$, we need to estimate the Sobol index of all sets of features $\kappa\in\mathcal{K}$ such that $k\in\kappa$. To minimize the computational cost of this computation, \cite{fel_look_2021} introduced an efficient sampling strategy based on Quasi-Monte Carlo methods \cite{morokoff_quasi-monte_1995} to generate the $N$ perturbations of dimension $K$ applied to the input and used Jansen's estimator \cite{jansen_analysis_1999} to estimate the TSIs given the models' outputs and the quasi-random perturbations. Their approach requires $N(K+2)$ forwards \cite{fel_look_2021}. 

To estimate the TSIs, they draw two matrices from a Quasi-Monte Carlo sequence of size $N\times K$ and convert them into perturbations, which they apply to $X$. The perturbated input yields two matrices, $A$ and $B$. $a_{jk}$ (resp. $b_{kj}$) is the element of $A$ (resp. $B$) corresponding to the $k^{th}$ feature and the $j^{th}$ sample. For the $k^{th}$ feature, they define $C^{(k)}$ in the same way as $A$, except that the column corresponding to feature $k$ is replaced by the column of $B$. They then derive an empirical estimator for the Sobol index and TSI as
\begin{equation}
    \hat{S}_k = \frac{\hat{V} - \frac{1}{2N}\sum_{j=1}^N \left[f(B_j) - f\left(C_j^{(k)}\right)\right]^2}{\hat{V}}, \;\;\;\;\;\;\; \hat{S}_{T_k} = \frac{\frac{1}{2N}\sum_{j=1}^N \left[f(A_j) - f\left(C_j^{(k)}\right)\right]^2}{\hat{V}},
\end{equation}
where $f_\emptyset = \displaystyle{
\frac{1}{N}\sum_{j=1}^N f(A_j)
}$ and $\hat{V} = \displaystyle{
\frac{1}{N-1}\sum_{j=0}^N \left[f\left(A_j\right) - f_\emptyset\right]^2
}$. Further implementational details can be found in \cite{fel_look_2021}.

\subsection{The wavelet scale attribution method (WCAM)}

\paragraph{Overview}

\begin{figure}[h]
\small
    \centering
    \includegraphics[width = \textwidth]{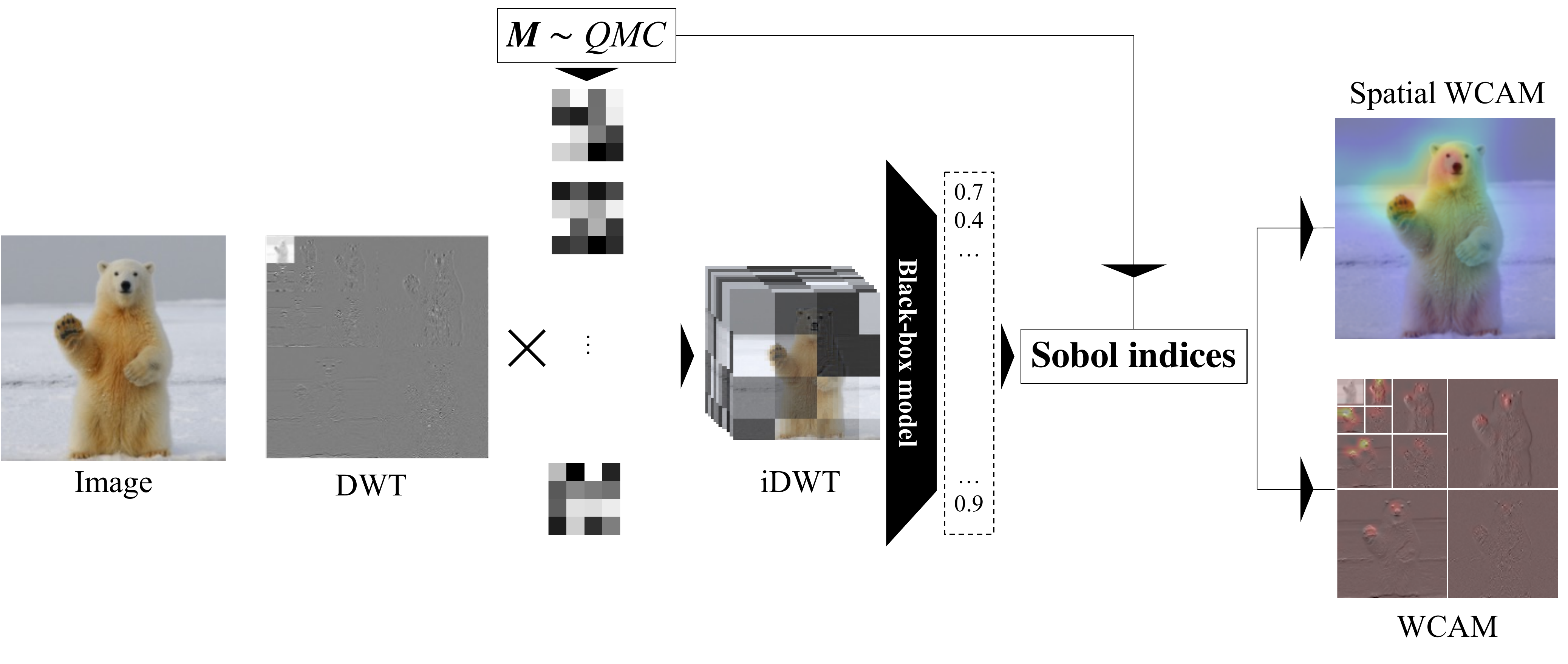}
    \caption{Flowchart of the wavelet scale attribution method (WCAM).
    }
    \label{fig:wcam-principle}
\end{figure}

The {\bf W}avelet s{\bf C}ale {\bf A}ttribution {\bf M}ethod ({\bf WCAM}) is an attribution method that quantifies the importance of the regions of the wavelet transform of an image to the predictions of a model. \autoref{fig:wcam-principle} depicts the principle of the WCAM. The importance of the regions of the wavelet transform of the input image is estimated by {\bf (1)} generating masks from a Quasi-Monte Carlo sequence, {\bf (2)} evaluating the model on perturbed images. We obtain these images by computing the DWT of the original image, applying the masks on the DWT to obtain perturbed DWT,\footnote{On an RGB image, we apply the DWT channel-wise and apply the same perturbation to each channel.} and inverting the perturbed DWT to generate perturbed images. We generate $N(K+2)$ perturbed images for a single image. {\bf (3)} We estimate the total Sobol indices of the perturbed regions of the wavelet transform using the masks and the model's outputs using Jansen's estimator \cite{jansen_analysis_1999}. \cite{fel_look_2021} introduced this approach to estimate the importance of image regions in the pixel space. We generalize it to the wavelet domain.

\begin{figure}[h]
    \centering
    \includegraphics[width = \textwidth]{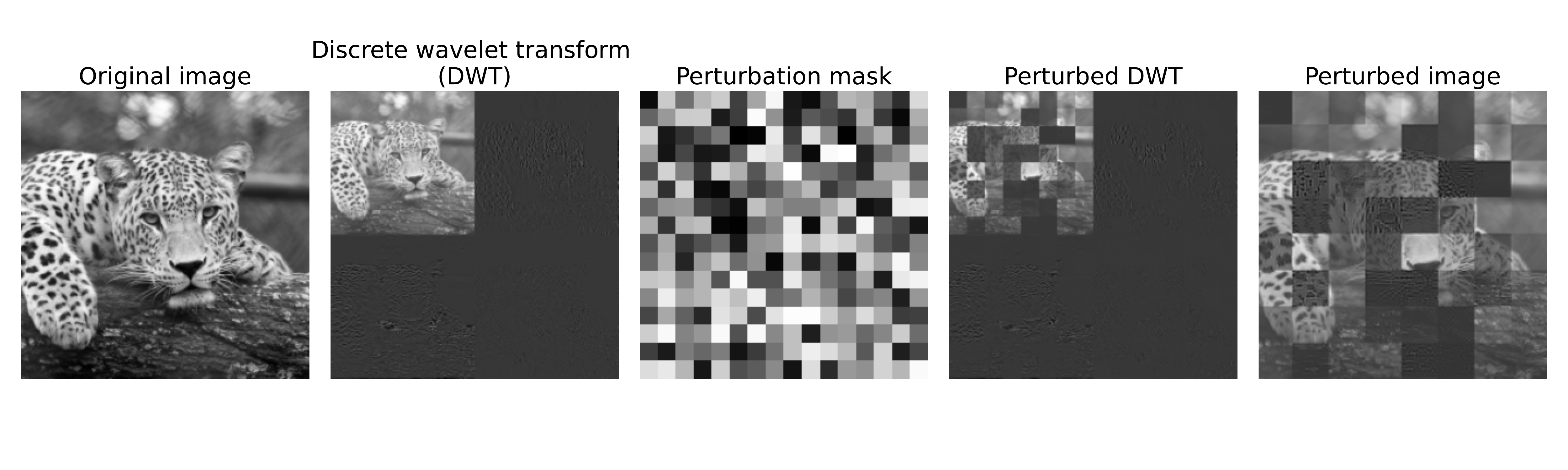}
    \caption{Workflow on a grayscale image and for a 1-level wavelet transform. We first compute the discrete wavelet transform of the image, then apply a mask on the DWT. It yields the perturbed DWT, which we invert to generate the perturbed image. We evaluate the model on the perturbed image.}
    \label{fig:wokflow-example}
\end{figure}

\paragraph{Generation of the masks} We follow the sampling procedure introduced by \cite{fel_look_2021} to generate the masks. Their approach involves drawing two independent matrices of size $N\times K$ from a Sobol low discrepancy sequence. $N$ is the number of designs necessary to estimate the variance, and $K$ is the sequence dimension. We reshape this sequence as a two-dimensional mask to generate our perturbation. By default, we perturb the wavelet transform with a mask of size 28 $\times$ 28 to balance between the dimensionality of the sequence and the accuracy of the perturbation. We reshape the 784-dimensional sequence to a grid of 28 $\times$ 28 to define our perturbation masks. We tried alternative mapping from the unidimensional sequence to the mask but with limited effect on the dimensionality reduction and at the expense of the meaningfulness of the perturbation in the wavelet domain. \autoref{fig:wokflow-example} illustrates our workflow for one mask to generate the images that are then passed to the model. 

\begin{figure}[h]
  \centering
  \includegraphics[width=0.7\linewidth]{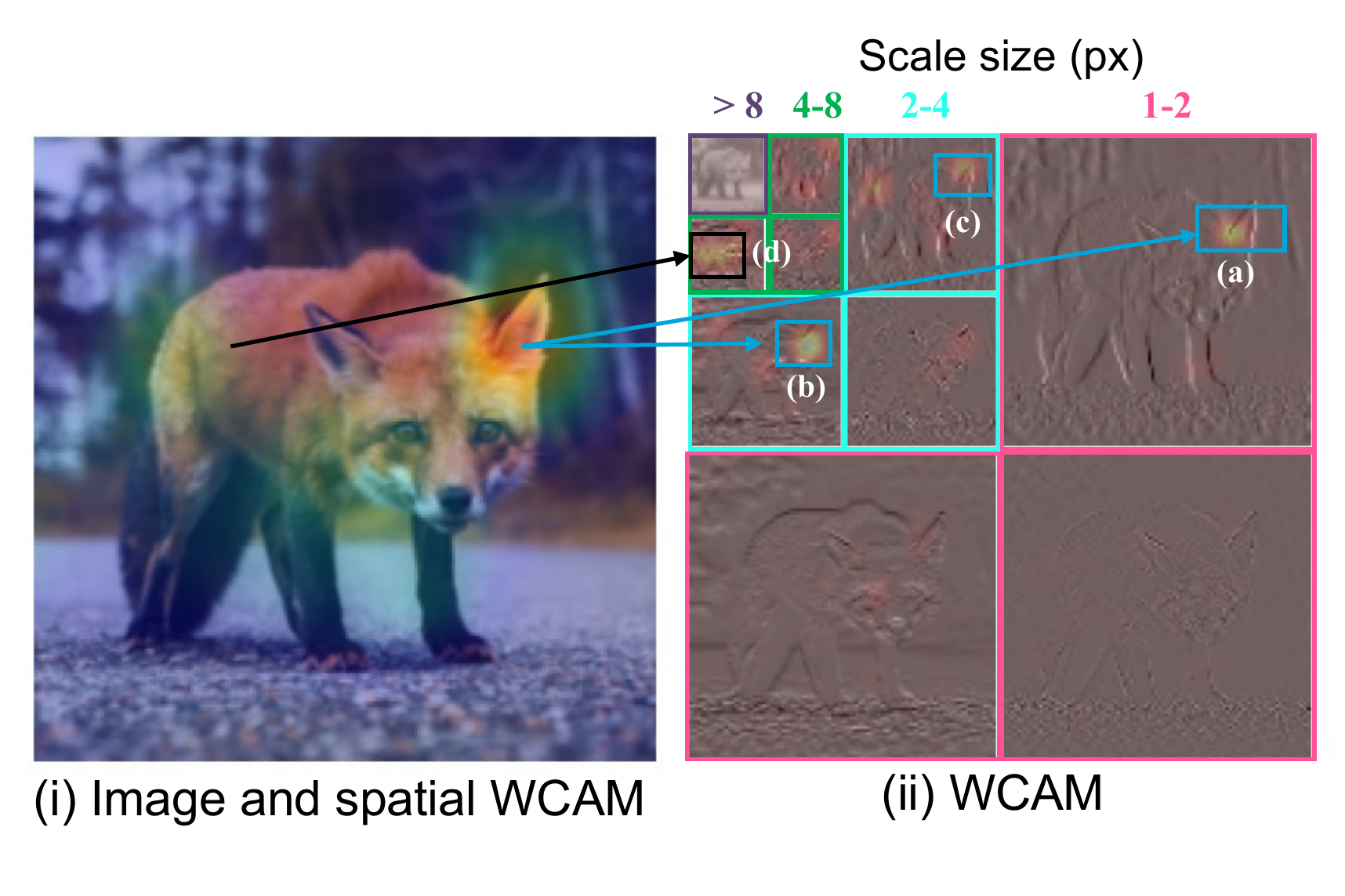}
  \caption{Decomposition of a prediction from the pixel domain (i) into the wavelet domain (ii) with the WCAM}
  \label{fig:ex-wcam-spatial-wcam}
\end{figure}

\paragraph{The WCAM expands attribution to the space-scale domain} The WCAM decomposes a prediction into the wavelet domain. As \autoref{fig:ex-wcam-spatial-wcam} depicts, highlighting an important area in the pixel domain (i) does not provide information on {\it what} the model sees. By decomposing the prediction into the wavelet domain (ii), the WCAM represents the important features of a prediction in terms of structural components. In the example of \autoref{fig:ex-wcam-spatial-wcam}, we can see two important areas for predicting the fox: the hind leg and the ear. We can see that three distinct components contribute to the prediction for the ear. Areas {\bf (a)}, {\bf (b)}, {\bf (c)} and {\bf (d)} highlight these components. {\bf (a)} corresponds to details at the 1-2 pixel scale, i.e., fine-grained details such as the fur in the ear. {\bf (b)} corresponds to details at the 2-4 pixel scale, i.e., larger details such as the shape of the ear. We can see that both vertical ({\bf (b)}) and horizontal ({\bf (c)}) components of the shape of the ear contribute to the prediction. On the other hand, for the hind leg, only the overall shape (4-8 pixel size, {\bf (d)}) contributes to the prediction. 


\section{Results and use cases}



\subsection{Results on evaluation benchmarks}

\paragraph{$\mu$-Fidelity} We evaluate our method the $\mu$-Fidelity, introduced by \cite{bhatt_evaluating_2020}. Contrary to insertion and deletion, which are area-under-curve metrics, the $\mu$-Fidelity is a correlation metric. It measures the correlation between the decrease of the predicted probabilities when features are in a baseline state and the importance of these features. We have
\begin{equation}
\mu\textrm{-Fidelity} = \underset{u \subseteq \{1, ..., K\},\atop |u| = d}{\operatorname{Corr}}\left( \sum_{i \in u} g(x_i)  , f(x) - f(x_{x_{u} = x_0})\right)
\end{equation}
where $g$ is the explanation function ({\it i.e.,} the explanation method), which quantifies the importance of the set of features $u$.

\paragraph{Results} In \autoref{tab:mu-fidelity}, we evaluate our method against a range of popular methods and across various model architectures. Results show that we outperform existing black-box methods and are competitive with white-box attribution methods. The projection in the space-scale domain is the cause for the superiority of our method: we can see that the WCAM shows that the coarser scales are essential for a prediction. When flattening the WCAM accross scales (see appendix \ref{sec:wcam-attribution} for more details) to derive the Spatial WCAM, our method performs similarly to other attribution methods. In appendix \ref{sec:wcam-attribution}, we provide additional evaluation results using the insertion and deletion scores \cite{petsiuk_rise_2018}.

\begin{table}[h]
\small
  \centering
  \vspace{0mm}\caption{{\bf $\mu$-Fidelity} scores obtained on 100 ImageNet validation set images (higher is better). The best results are \textbf{bolded} and second best \underline{underlined}. All benchmarks use the Xplique library \cite{fel_xplique_2022}.
  }\label{tab:mu-fidelity}
  \resizebox{\textwidth}{!}{\begin{tabular}{c lcccc}
  \toprule
   
   & Method & \textit{VGG16} \cite{simonyan_very_2015} & \textit{ResNet50} \cite{he_deep_2016} & \textit{MobileNet} \cite{howard_mobilenets_2017} & \textit{EfficientNet} \cite{tan_efficientnet_2020} \\

    \midrule
   $\mu$-Fidelity ($\uparrow$)  & & & & & \\

  \multirow{5}{*}{White-box}
    & Saliency~\cite{simonyan_deep_2014} &  0.043  & 0.060  & -0.002  & 0.052  \\ 
  & Grad.-Input~\cite{shrikumar_not_2017} &\underline{0.105}  & 0.051  & 0.023  & 0.030  \\
  & Integ.-Grad.~\cite{sundararajan_axiomatic_2017} & {\bf 0.137}  & {\bf 0.112}  & \underline{0.130}  & {\bf 0.134} \\
  & GradCAM++~\cite{selvaraju_grad-cam_2020} &  0.089  & 0.083  & -0.001  & 0.063   \\ 
  & VarGrad~\cite{selvaraju_grad-cam_2020} & 0.054  & \underline{0.099}  &{\bf  0.279}  & \underline{0.093}  \\

  \midrule  
  \multirow{3}{*}{Black-box}
  & RISE~\cite{petsiuk_rise_2018} & 0.020  & 0.074 & -0.025  & \underline{0.042}  \\ 
  & Sobol \cite{fel_look_2021} &0.095 & \underline{ 0.108}  & -0.036  & 0.013  \\ 
  & Spatial WCAM (ours)  &0.016  & -0.037 & \underline{0.020} & -0.016\\ 
  &  WCAM (ours)  & {\bf 0.197}  & {\bf 0.191} & {\bf 0.105}  & {\bf 0.187}\\ 
  \bottomrule
  \end{tabular}}
  \end{table}

\subsection{Assessing the robustness of a prediction}

\begin{figure}[h]
\centering
  \centering
  \includegraphics[width=.6\textwidth]{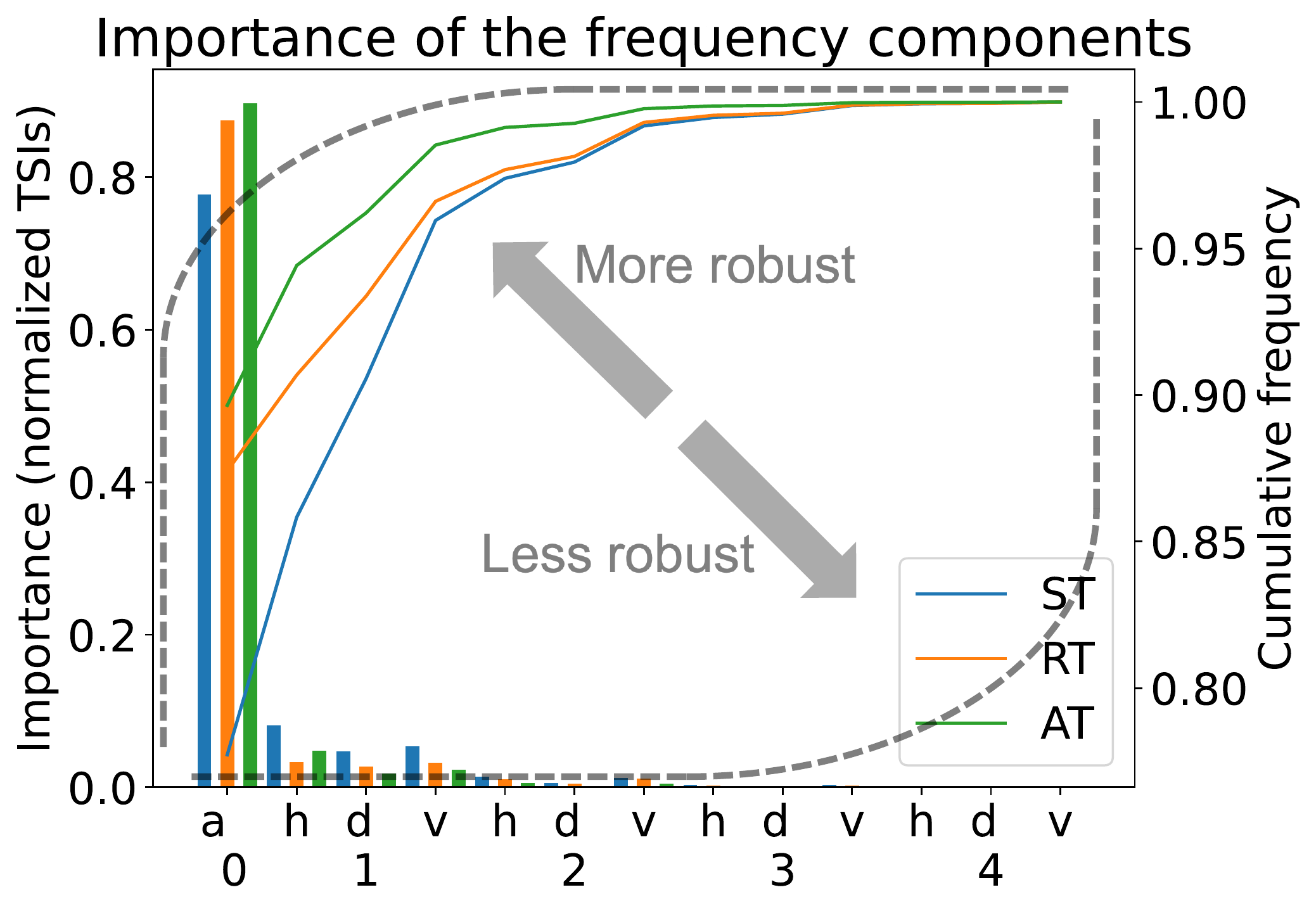}
  \label{fig:approx}
\caption{
Representation of the scales of the WCAM as frequencies. Levels (numbered from (0 to 4) indicate the scales, from the coarser (i.e., lowest frequencies) to the finest (i.e., highest frequencies. The level 0 or "a" corresponds to the approximation coefficients. Labels "h," "v," and "d" correspond to the horizontal, vertical, and diagonal details, respectively. The rightmost index plots the cumulative curve. "AT," "RT," and "ST" stand for adversarial, robust, and standard training, respectively.
}

    \label{fig:fcam-wcam-equivalence}
\end{figure}

\paragraph{Scales, frequencies, and robustness} Scales in the wavelet domain correspond to dyadic frequency ranges in the Fourier domain. The smallest scales correspond to the highest frequencies. Therefore, the WCAM connects attribution with frequency-centric approaches to model robustness. \autoref{fig:fcam-wcam-equivalence} uses the WCAM to quantify the model's robustness. Following \cite{chen_rethinking_2022}, we distinguish models trained with adversarial training ("AT," \cite{madry_towards_2019,qu_fast_2017,shafahi_adversarial_2019}), robust training ("RT," \cite{hendrycks_augmix_2020,hendrycks_pixmix_2022,geirhos_imagenet-trained_2023}) and standard training ("ST," e.g., ERM \cite{vapnik_nature_1999}). We can see that AT and RT models favor coarse scales (i.e., low-frequencies) over fine scales (i.e., high-frequencies). The WCAM characterizes robust models by estimating the importance of each frequency component in the final prediction. We can see that the ordering from the detail coefficients corresponding to the largest scales from those corresponding to the highest remains the same. This can be further assessed by the cumulative frequencies (right axis), where the most robust models have a more concentrated cumulative curve than the least robust models. These results are in line with existing works \cite{zhang_range_2022,chen_rethinking_2022,wang_high-frequency_2020,yin_fourier_2020} and show that the WCAM correctly estimates the robustness of a model.

\paragraph{Relevance of the important scales in expert settings}

The wavelet transform provides a multi-resolution decomposition of an image \cite{mallat_wavelet_1999}. In other words, it isolates textures, edges, and shapes with respect to their scales and locations. The WCAM lets us see whether a model relies more on textures or shapes, which is impossible with existing attribution methods. In many settings, such information is valuable as it is interpreted in terms of specific features of the object of interest. Therefore, the WCAM links visual features (the model relies on) and semantic features (from which humans draw conclusions). For instance, in brain tumor classification from magnetic resonance imaging (MRI), small-scale details correspond to interpapillary capillary loop patterns \cite{garcia-peraza-herrera_intrapapillary_2020}. In natural sciences, small scales can correspond to veins or small patterns on specific species of leaves \cite{wilf_computer_2016,spagnuolo_decoding_2022}. Finally, in remote sensing, scales correspond to different structural properties of small objects \cite{kasmi_crowdsourced_2023,kasmi_can_2023}, and information in the finer scales is more sensitive to image quality. If the latter varies too much (e.g., due to different acquisition conditions and signal-to-noise ratios), then information at these scales may disappear.

\paragraph{Towards assessing the reliability of a prediction} We argue that a reliable prediction should be both relevant and robust. It is relevant as it relies on expected factors (or more generally on non-spurious factors) and robust as we want our prediction to be invariant to perturbations that can occur during the data acquisition process (e.g., heterogeneous properties of the optical transfer function or alterations of the signal-to-noise ratio) and limit the generalization capabilities of the model. Some relevant factors may be unreliable: in this case, the practitioner needs to be aware of that and adjust the data acquisition process. On the other hand, some models that are provably more robust may not be reliable if they rely on spurious components. The WCAM is a first step towards increasing the reliability of deep models in real-life settings as it unveils what scales are important and whether they are robust or not.








\section{Conclusions and future work}

We introduce the {\bf W}avelet s{\bf C}ale {\bf A}ttribution {\bf M}ethod ({\bf WCAM}), a generalization attribution to the space-scale domain. The WCAM highlights the important regions in the space-scale domain using efficient perturbations of the wavelet transform of the input image. We estimate the contribution of the regions of the wavelet transform using total Sobol indices. Compared to existing attribution methods, the WCAM identifies what scales, which correspond to semantic features, are important for a prediction, thus providing more guidance regarding the relevance of the decision process of a model. Moreover, the WCAM connects attribution with robustness as scales in the wavelet domain correspond to frequencies in the Fourier domain. Applications of the WCAM lie in expert settings where practitioners need to evaluate the reliability of the prediction made by the model: does it rely on {\it relevant} components, and are these components {\it robust} to input perturbations that can occur in real life? Such assessment is crucial to improve the trustworthiness of deep learning models, especially as these models seem to be scale-inconsistent and rely on factors for contextual rather than semantical reasons. 

\paragraph{Limitations and future works} The main limitation of our approach is that it is computationally more expensive than existing black-box attribution methods. We plan to evaluate the benefits of the WCAM in an expert setting: the remote sensing of rooftop photovoltaic (PV) installations, where current models lack reliability due to their sensitivity to acquisition conditions \cite{kasmi_crowdsourced_2023}.



\section{Acknowledgements}

This work is funded by RTE France, the French transmission system operator, and benefited from CIFRE funding from the ANRT. The authors gratefully acknowledge the support of this project. The authors would like to thank Thomas Fel, the first author of the original article {\it "Look at the Variance! Efficient Black-box Explanations with Sobol-based Sensitivity Analysis"} for his insightful comments and advice while preparing this work. The authors would also like to thank Hugo Thimonier for his helpful advice and feedback and Thomas Heggarty for proofreading the final manuscript.

{\small
\bibliographystyle{plain}
\bibliography{references}

\begin{thebibliography}{10}

\bibitem{achtibat_where_2022}
Reduan Achtibat, Maximilian Dreyer, Ilona Eisenbraun, Sebastian Bosse, Thomas
  Wiegand, Wojciech Samek, and Sebastian Lapuschkin.
\newblock From "{Where}" to "{What}": {Towards} {Human}-{Understandable}
  {Explanations} through {Concept} {Relevance} {Propagation}, June 2022.
\newblock arXiv:2206.03208 [cs].

\bibitem{bhatt_evaluating_2020}
Umang Bhatt, Adrian Weller, and José M.~F. Moura.
\newblock Evaluating and {Aggregating} {Feature}-based {Model} {Explanations},
  May 2020.
\newblock arXiv:2005.00631 [cs, stat].

\bibitem{carvalho_machine_2019}
Diogo~V. Carvalho, Eduardo~M. Pereira, and Jaime~S. Cardoso.
\newblock Machine {Learning} {Interpretability}: {A} {Survey} on {Methods} and
  {Metrics}.
\newblock {\em Electronics}, 8(8):832, July 2019.

\bibitem{chen_rethinking_2022}
Yiting Chen, Qibing Ren, and Junchi Yan.
\newblock Rethinking and {Improving} {Robustness} of {Convolutional} {Neural}
  {Networks}: a {Shapley} {Value}-based {Approach} in {Frequency} {Domain}.
\newblock October 2022.

\bibitem{delseny_white_2021}
Hervé Delseny, Christophe Gabreau, Adrien Gauffriau, Bernard Beaudouin,
  Ludovic Ponsolle, Lucian Alecu, Hugues Bonnin, Brice Beltran, Didier Duchel,
  Jean-Brice Ginestet, Alexandre Hervieu, Ghilaine Martinez, Sylvain Pasquet,
  Kevin Delmas, Claire Pagetti, Jean-Marc Gabriel, Camille Chapdelaine,
  Sylvaine Picard, Mathieu Damour, Cyril Cappi, Laurent Gardès, Florence
  De~Grancey, Eric Jenn, Baptiste Lefevre, Gregory Flandin, Sébastien
  Gerchinovitz, Franck Mamalet, and Alexandre Albore.
\newblock White {Paper} {Machine} {Learning} in {Certified} {Systems}, March
  2021.
\newblock arXiv:2103.10529 [cs] version: 1.

\bibitem{dhar_modeling_2018}
Manik Dhar, Aditya Grover, and Stefano Ermon.
\newblock Modeling {Sparse} {Deviations} for {Compressed} {Sensing} using
  {Generative} {Models}, July 2018.
\newblock arXiv:1807.01442 [cs, stat].

\bibitem{fel_look_2021}
Thomas Fel, Remi Cadene, Mathieu Chalvidal, Matthieu Cord, David Vigouroux, and
  Thomas Serre.
\newblock Look at the {Variance}! {Efficient} {Black}-box {Explanations} with
  {Sobol}-based {Sensitivity} {Analysis}, November 2021.
\newblock arXiv:2111.04138 [cs].

\bibitem{fel_dont_2023}
Thomas Fel, Melanie Ducoffe, David Vigouroux, Remi Cadene, Mikael Capelle,
  Claire Nicodeme, and Thomas Serre.
\newblock Don't {Lie} to {Me}! {Robust} and {Efficient} {Explainability} with
  {Verified} {Perturbation} {Analysis}, March 2023.
\newblock arXiv:2202.07728 [cs].

\bibitem{fel_xplique_2022}
Thomas Fel, Lucas Hervier, David Vigouroux, Antonin Poche, Justin Plakoo, Remi
  Cadene, Mathieu Chalvidal, Julien Colin, Thibaut Boissin, Louis Bethune,
  Agustin Picard, Claire Nicodeme, Laurent Gardes, Gregory Flandin, and Thomas
  Serre.
\newblock Xplique: {A} {Deep} {Learning} {Explainability} {Toolbox}, June 2022.
\newblock arXiv:2206.04394 [cs].

\bibitem{fel_craft_2023}
Thomas Fel, Agustin Picard, Louis Bethune, Thibaut Boissin, David Vigouroux,
  Julien Colin, Rémi Cadène, and Thomas Serre.
\newblock {CRAFT}: {Concept} {Recursive} {Activation} {FacTorization} for
  {Explainability}, March 2023.
\newblock arXiv:2211.10154 [cs].

\bibitem{garcia-peraza-herrera_intrapapillary_2020}
Luis~C. García-Peraza-Herrera, Martin Everson, Laurence Lovat, Hsiu-Po Wang,
  Wen~Lun Wang, Rehan Haidry, Danail Stoyanov, Sébastien Ourselin, and Tom
  Vercauteren.
\newblock Intrapapillary capillary loop classification in magnification
  endoscopy: open dataset and baseline methodology.
\newblock {\em International Journal of Computer Assisted Radiology and
  Surgery}, 15(4):651--659, April 2020.

\bibitem{geirhos_shortcut_2020}
Robert Geirhos, Jörn-Henrik Jacobsen, Claudio Michaelis, Richard Zemel,
  Wieland Brendel, Matthias Bethge, and Felix~A. Wichmann.
\newblock Shortcut {Learning} in {Deep} {Neural} {Networks}.
\newblock {\em Nature Machine Intelligence}, 2(11):665--673, November 2020.
\newblock arXiv:2004.07780 [cs, q-bio].

\bibitem{geirhos_imagenet-trained_2023}
Robert Geirhos, Patricia Rubisch, Claudio Michaelis, Matthias Bethge, Felix~A.
  Wichmann, and Wieland Brendel.
\newblock {ImageNet}-trained {CNNs} are biased towards texture; increasing
  shape bias improves accuracy and robustness.
\newblock April 2023.

\bibitem{hagras_toward_2018}
Hani Hagras.
\newblock Toward {Human}-{Understandable}, {Explainable} {AI}.
\newblock {\em Computer}, 51(9):28--36, September 2018.
\newblock Conference Name: Computer.

\bibitem{halton_efficiency_1960}
J.~H. Halton.
\newblock On the efficiency of certain quasi-random sequences of points in
  evaluating multi-dimensional integrals.
\newblock {\em Numerische Mathematik}, 2(1):84--90, December 1960.

\bibitem{he_deep_2016}
Kaiming He, Xiangyu Zhang, Shaoqing Ren, and Jian Sun.
\newblock Deep residual learning for image recognition.
\newblock In {\em Proceedings of the {IEEE} conference on computer vision and
  pattern recognition}, pages 770--778, 2016.

\bibitem{hendrycks_benchmarking_2019}
Dan Hendrycks and Thomas Dietterich.
\newblock Benchmarking {Neural} {Network} {Robustness} to {Common}
  {Corruptions} and {Perturbations}, March 2019.
\newblock arXiv:1903.12261 [cs, stat].

\bibitem{hendrycks_augmix_2020}
Dan Hendrycks, Norman Mu, Ekin~D. Cubuk, Barret Zoph, Justin Gilmer, and Balaji
  Lakshminarayanan.
\newblock {AugMix}: {A} {Simple} {Data} {Processing} {Method} to {Improve}
  {Robustness} and {Uncertainty}, February 2020.
\newblock arXiv:1912.02781 [cs, stat].

\bibitem{hendrycks_pixmix_2022}
Dan Hendrycks, Andy Zou, Mantas Mazeika, Leonard Tang, Bo~Li, Dawn Song, and
  Jacob Steinhardt.
\newblock {PixMix}: {Dreamlike} {Pictures} {Comprehensively} {Improve} {Safety}
  {Measures}, March 2022.
\newblock arXiv:2112.05135 [cs].

\bibitem{hoeffding_class_1992}
Wassily Hoeffding.
\newblock A {Class} of {Statistics} with {Asymptotically} {Normal}
  {Distribution}.
\newblock In Samuel Kotz and Norman~L. Johnson, editors, {\em Breakthroughs in
  {Statistics}: {Foundations} and {Basic} {Theory}}, Springer {Series} in
  {Statistics}, pages 308--334. Springer, New York, NY, 1992.

\bibitem{homma_importance_1996}
Toshimitsu Homma and Andrea Saltelli.
\newblock Importance measures in global sensitivity analysis of nonlinear
  models.
\newblock {\em Reliability Engineering \& System Safety}, 52(1):1--17, April
  1996.

\bibitem{howard_mobilenets_2017}
Andrew~G. Howard, Menglong Zhu, Bo~Chen, Dmitry Kalenichenko, Weijun Wang,
  Tobias Weyand, Marco Andreetto, and Hartwig Adam.
\newblock {MobileNets}: {Efficient} {Convolutional} {Neural} {Networks} for
  {Mobile} {Vision} {Applications}, April 2017.
\newblock arXiv:1704.04861 [cs].

\bibitem{jansen_analysis_1999}
Michiel J.~W. Jansen.
\newblock Analysis of variance designs for model output.
\newblock {\em Computer Physics Communications}, 117(1):35--43, March 1999.

\bibitem{jo_measuring_2017}
Jason Jo and Yoshua Bengio.
\newblock Measuring the tendency of {CNNs} to {Learn} {Surface} {Statistical}
  {Regularities}, November 2017.
\newblock arXiv:1711.11561 [cs, stat].

\bibitem{kasmi_can_2023}
Gabriel Kasmi, Laurent Dubus, Yves-Marie Saint-Drenan, and Philippe Blanc.
\newblock Can {We} {Reliably} {Improve} the {Robustness} to {Image}
  {Acquisition} of {Remote} {Sensing} of {PV} {Systems}?, September 2023.
\newblock arXiv:2309.12214 [cs].

\bibitem{kasmi_crowdsourced_2023}
Gabriel Kasmi, Yves-Marie Saint-Drenan, David Trebosc, Raphaël Jolivet,
  Jonathan Leloux, Babacar Sarr, and Laurent Dubus.
\newblock A crowdsourced dataset of aerial images with annotated solar
  photovoltaic arrays and installation metadata.
\newblock {\em Scientific Data}, 10(1):59, January 2023.

\bibitem{koh_understanding_2020}
Pang~Wei Koh and Percy Liang.
\newblock Understanding {Black}-box {Predictions} via {Influence} {Functions},
  December 2020.
\newblock arXiv:1703.04730 [cs, stat].

\bibitem{koh_wilds_2021}
Pang~Wei Koh, Shiori Sagawa, Henrik Marklund, Sang~Michael Xie, Marvin Zhang,
  Akshay Balsubramani, Weihua Hu, Michihiro Yasunaga, Richard~Lanas Phillips,
  Irena Gao, and {others}.
\newblock Wilds: {A} benchmark of in-the-wild distribution shifts.
\newblock In {\em International {Conference} on {Machine} {Learning}}, pages
  5637--5664. PMLR, 2021.

\bibitem{lundberg_unified_2017}
Scott~M Lundberg and Su-In Lee.
\newblock A unified approach to interpreting model predictions.
\newblock {\em Advances in neural information processing systems}, 30, 2017.

\bibitem{madry_towards_2019}
Aleksander Madry, Aleksandar Makelov, Ludwig Schmidt, Dimitris Tsipras, and
  Adrian Vladu.
\newblock Towards {Deep} {Learning} {Models} {Resistant} to {Adversarial}
  {Attacks}, September 2019.
\newblock arXiv:1706.06083 [cs, stat].

\bibitem{mallat_theory_1989}
S.G. Mallat.
\newblock A theory for multiresolution signal decomposition: the wavelet
  representation.
\newblock {\em IEEE Transactions on Pattern Analysis and Machine Intelligence},
  11(7):674--693, July 1989.
\newblock Conference Name: IEEE Transactions on Pattern Analysis and Machine
  Intelligence.

\bibitem{mallat_wavelet_1999}
Stéphane Mallat.
\newblock {\em A wavelet tour of signal processing}.
\newblock Elsevier, 1999.

\bibitem{mckay_comparison_1979}
M.~D. McKay, R.~J. Beckman, and W.~J. Conover.
\newblock A {Comparison} of {Three} {Methods} for {Selecting} {Values} of
  {Input} {Variables} in the {Analysis} of {Output} from a {Computer} {Code}.
\newblock {\em Technometrics}, 21(2):239--245, 1979.
\newblock Publisher: [Taylor \& Francis, Ltd., American Statistical
  Association, American Society for Quality].

\bibitem{morokoff_quasi-monte_1995}
William~J. Morokoff and Russel~E. Caflisch.
\newblock Quasi-{Monte} {Carlo} {Integration}.
\newblock {\em Journal of Computational Physics}, 122(2):218--230, December
  1995.

\bibitem{novello_making_2022}
Paul Novello, Thomas Fel, and David Vigouroux.
\newblock Making {Sense} of {Dependence}: {Efficient} {Black}-box
  {Explanations} {Using} {Dependence} {Measure}, September 2022.
\newblock arXiv:2206.06219 [cs, stat].

\bibitem{petsiuk_rise_2018}
Vitali Petsiuk, Abir Das, and Kate Saenko.
\newblock {RISE}: {Randomized} {Input} {Sampling} for {Explanation} of
  {Black}-box {Models}, September 2018.
\newblock arXiv:1806.07421 [cs].

\bibitem{pezeshki_gradient_2021}
Mohammad Pezeshki, Sékou-Oumar Kaba, Yoshua Bengio, Aaron Courville, Doina
  Precup, and Guillaume Lajoie.
\newblock Gradient {Starvation}: {A} {Learning} {Proclivity} in {Neural}
  {Networks}, November 2021.
\newblock arXiv:2011.09468 [cs, math, stat].

\bibitem{puli_dont_2023}
Aahlad Puli, Lily Zhang, Yoav Wald, and Rajesh Ranganath.
\newblock Don't blame {Dataset} {Shift}! {Shortcut} {Learning} due to
  {Gradients} and {Cross} {Entropy}, August 2023.
\newblock arXiv:2308.12553 [cs, stat].

\bibitem{qu_fast_2017}
Zhipeng Qu, Armel Oumbe, Philippe Blanc, Bella Espinar, Gerhard Gesell, Benoît
  Gschwind, Lars Klüser, Mireille Lefèvre, Laurent Saboret, Marion
  Schroedter-Homscheidt, and Lucien Wald.
\newblock Fast radiative transfer parameterisation for assessing the surface
  solar irradiance: {The} {Heliosat}‑4 method.
\newblock {\em Meteorologische Zeitschrift}, 26(1):33--57, February 2017.

\bibitem{rahaman_spectral_2019}
Nasim Rahaman, Aristide Baratin, Devansh Arpit, Felix Draxler, Min Lin, Fred~A.
  Hamprecht, Yoshua Bengio, and Aaron Courville.
\newblock On the {Spectral} {Bias} of {Neural} {Networks}, May 2019.
\newblock arXiv:1806.08734 [cs, stat].

\bibitem{ribeiro_why_2016}
Marco~Tulio Ribeiro, Sameer Singh, and Carlos Guestrin.
\newblock "{Why} {Should} {I} {Trust} {You}?": {Explaining} the {Predictions}
  of {Any} {Classifier}, August 2016.
\newblock arXiv:1602.04938 [cs, stat].

\bibitem{russakovsky_imagenet_2015}
Olga Russakovsky, Jia Deng, Hao Su, Jonathan Krause, Sanjeev Satheesh, Sean Ma,
  Zhiheng Huang, Andrej Karpathy, Aditya Khosla, Michael Bernstein,
  Alexander~C. Berg, and Li~Fei-Fei.
\newblock {ImageNet} {Large} {Scale} {Visual} {Recognition} {Challenge},
  January 2015.
\newblock arXiv:1409.0575 [cs].

\bibitem{saria_tutorial_2019}
Suchi Saria and Adarsh Subbaswamy.
\newblock Tutorial: {Safe} and {Reliable} {Machine} {Learning}, April 2019.
\newblock arXiv:1904.07204 [cs].

\bibitem{selvaraju_grad-cam_2020}
Ramprasaath~R. Selvaraju, Michael Cogswell, Abhishek Das, Ramakrishna Vedantam,
  Devi Parikh, and Dhruv Batra.
\newblock Grad-{CAM}: {Visual} {Explanations} from {Deep} {Networks} via
  {Gradient}-based {Localization}.
\newblock {\em International Journal of Computer Vision}, 128(2):336--359,
  February 2020.
\newblock arXiv:1610.02391 [cs].

\bibitem{shafahi_adversarial_2019}
Ali Shafahi, Mahyar Najibi, Amin Ghiasi, Zheng Xu, John Dickerson, Christoph
  Studer, Larry~S. Davis, Gavin Taylor, and Tom Goldstein.
\newblock Adversarial {Training} for {Free}!, November 2019.
\newblock arXiv:1904.12843 [cs, stat].

\bibitem{shrikumar_not_2017}
Avanti Shrikumar, Peyton Greenside, Anna Shcherbina, and Anshul Kundaje.
\newblock Not {Just} a {Black} {Box}: {Learning} {Important} {Features}
  {Through} {Propagating} {Activation} {Differences}, April 2017.
\newblock arXiv:1605.01713 [cs].

\bibitem{simonyan_deep_2014}
Karen Simonyan, Andrea Vedaldi, and Andrew Zisserman.
\newblock Deep {Inside} {Convolutional} {Networks}: {Visualising} {Image}
  {Classification} {Models} and {Saliency} {Maps}, April 2014.
\newblock arXiv:1312.6034 [cs].

\bibitem{simonyan_very_2015}
Karen Simonyan and Andrew Zisserman.
\newblock Very {Deep} {Convolutional} {Networks} for {Large}-{Scale} {Image}
  {Recognition}, April 2015.
\newblock arXiv:1409.1556 [cs].

\bibitem{sobol_distribution_1967}
I.~M Sobol'.
\newblock On the distribution of points in a cube and the approximate
  evaluation of integrals.
\newblock {\em USSR Computational Mathematics and Mathematical Physics},
  7(4):86--112, January 1967.

\bibitem{sobol_sensitivity_1990}
Ilya~Meerovich Sobol.
\newblock On sensitivity estimation for nonlinear mathematical models.
\newblock {\em Matematicheskoe modelirovanie}, 2(1):112--118, 1990.
\newblock Publisher: Russian Academy of Sciences, Branch of Mathematical
  Sciences.

\bibitem{spagnuolo_decoding_2022}
Edward~J. Spagnuolo, Peter Wilf, and Thomas Serre.
\newblock Decoding family-level features for modern and fossil leaves from
  computer-vision heat maps.
\newblock {\em American Journal of Botany}, 109(5):768--788, 2022.
\newblock \_eprint: https://onlinelibrary.wiley.com/doi/pdf/10.1002/ajb2.1842.

\bibitem{sundararajan_axiomatic_2017}
Mukund Sundararajan, Ankur Taly, and Qiqi Yan.
\newblock Axiomatic {Attribution} for {Deep} {Networks}, June 2017.
\newblock arXiv:1703.01365 [cs].

\bibitem{tan_efficientnet_2020}
Mingxing Tan and Quoc~V. Le.
\newblock {EfficientNet}: {Rethinking} {Model} {Scaling} for {Convolutional}
  {Neural} {Networks}, September 2020.
\newblock arXiv:1905.11946 [cs, stat].

\bibitem{vapnik_nature_1999}
Vladimir Vapnik.
\newblock {\em The nature of statistical learning theory}.
\newblock Springer science \& business media, 1999.

\bibitem{wang_high-frequency_2020}
Haohan Wang, Xindi Wu, Zeyi Huang, and Eric~P. Xing.
\newblock High-{Frequency} {Component} {Helps} {Explain} the {Generalization}
  of {Convolutional} {Neural} {Networks}.
\newblock In {\em 2020 {IEEE}/{CVF} {Conference} on {Computer} {Vision} and
  {Pattern} {Recognition} ({CVPR})}, pages 8681--8691, Seattle, WA, USA, June
  2020. IEEE.

\bibitem{wang_frequency_2022}
Shunxin Wang, Raymond Veldhuis, Christoph Brune, and Nicola Strisciuglio.
\newblock Frequency {Shortcut} {Learning} in {Neural} {Networks}.
\newblock 2022.

\bibitem{wang_what_2023}
Shunxin Wang, Raymond Veldhuis, Christoph Brune, and Nicola Strisciuglio.
\newblock What do neural networks learn in image classification? {A} frequency
  shortcut perspective, July 2023.
\newblock arXiv:2307.09829 [cs].

\bibitem{wilf_computer_2016}
Peter Wilf, Shengping Zhang, Sharat Chikkerur, Stefan~A. Little, Scott~L. Wing,
  and Thomas Serre.
\newblock Computer vision cracks the leaf code.
\newblock {\em Proceedings of the National Academy of Sciences of the United
  States of America}, 113(12):3305--3310, March 2016.

\bibitem{xu_frequency_2020}
Zhi-Qin~John Xu, Yaoyu Zhang, Tao Luo, Yanyang Xiao, and Zheng Ma.
\newblock Frequency {Principle}: {Fourier} {Analysis} {Sheds} {Light} on {Deep}
  {Neural} {Networks}.
\newblock {\em Communications in Computational Physics}, 28(5):1746--1767, June
  2020.
\newblock arXiv:1901.06523 [cs, stat].

\bibitem{yin_fourier_2020}
Dong Yin, Raphael~Gontijo Lopes, Jonathon Shlens, Ekin~D. Cubuk, and Justin
  Gilmer.
\newblock A {Fourier} {Perspective} on {Model} {Robustness} in {Computer}
  {Vision}, September 2020.
\newblock arXiv:1906.08988 [cs, stat].

\bibitem{zeiler_visualizing_2013}
Matthew~D. Zeiler and Rob Fergus.
\newblock Visualizing and {Understanding} {Convolutional} {Networks}, November
  2013.
\newblock arXiv:1311.2901 [cs].

\bibitem{zhang_understanding_2016}
Chiyuan Zhang, Samy Bengio, Moritz Hardt, Benjamin Recht, and Oriol Vinyals.
\newblock Understanding deep learning requires rethinking generalization.
\newblock {\em arXiv preprint arXiv:1611.03530}, 2016.

\bibitem{zhang_range_2022}
Zhuang Zhang, Dejian Meng, Lijun Zhang, Wei Xiao, and Wei Tian.
\newblock The range of harmful frequency for {DNN} corruption robustness.
\newblock {\em Neurocomputing}, 481:294--309, April 2022.

\end{thebibliography}
}
\newpage
\appendix
\section{Implementational details on the WCAM}\label{sec:wcam-appendix}




\subsection{Sensitivity analysis}

\subsubsection{Effect of the grid size, number of designs and sampler on the explanations}\label{sec:parameters}

To evaluate the effect of the parameters {\tt grid\_size} and {\tt nb\_design} and of the samplers on the estimation of the Sobol indices, we consider 100 images sampled from ImageNet and compute the Insertion \cite{petsiuk_rise_2018} and Deletion \cite{petsiuk_rise_2018} scores. These pointing metrics measure the accuracy of an explanation. We compute these scores on the {\it spatial}  WCAM to study the effect of the parameters. Benchmarks were carried out using the Xplique toolbox \cite{fel_xplique_2022}.

\paragraph{Evaluation metrics} We evaluate the accuracy of our explanations based on two metrics introduced by \cite{petsiuk_rise_2018}. The first one is deletion. Deletion measures the drop in probability in the predicted probability of a class when removing the pixels highlighted by the explanation. The higher the drop, the better the explanation. At step $t$, the $u$ most important variables according to deletion are given by

\begin{equation}
    \textrm{Del}^{(t)} = f\left(x_{x_u = x_0}\right),
\end{equation}
where $x_0$  is the baseline set, set to 0 in the Xplique library. 

On the other hand, the insertion measures the contribution of the pixels highlighted by the explanation to the predicted probability when {\it inserting} a feature. The higher the increase, the better the explanation. At step $t$, the insertion score for the $u$ most important features is given by

\begin{equation}
    \textrm{Ins}^{(t)} = f\left(x_{x_{\bar u} = x_0}\right),
\end{equation}

where $x_0$ is the same baseline state as deletion. Insertion and deletion are area-under-curve metrics; the lower the deletion, the better, and the better the insertion. The intuition is as follows: if an explanation picks the most important features (according to the probability score), we should remove (resp. insert) the most important feature at step $t=1$, the second most important at step $t=2$, etc. Therefore, the probability drop (resp. increase) is the highest at the first step and gradually decreases. 

\paragraph{Parameters} The {\tt grid\_size} parameter defines how coarse the explanation will be. On the wavelet transform, each pixel corresponds to a wavelet coefficient. Therefore, a grid size that matches the input size will explain all wavelet coefficients individually. However, the computational cost of such an explanation is prohibitive. Therefore, we explain {\it sets} of wavelet coefficients as the grid size is lower than the input size. To correctly explore the variability in space, our grid size should not be too large. After empirical investigation, we chose a grid size of 28 $\times$ 28 on a 224 $\times$ 224 input image. Additionally, we tested values ranging from 8 to 32, 8 corresponding to the default value of the Sobol Attribution Method \cite{fel_look_2021} and 32 allowing for a 4-level decomposition. 

The parameter {\tt nb\_design} corresponds to the number of deviations from the means used to estimate the variance of the Sobol indices. Theoretically, this parameter should be as high as possible for an accurate variance estimation. In practice, this increases the number of forwards and, hence, the computation time of the total Sobol indices. As the implementation of the spectral attribution requires more operations than the Sobol attributions, we want to keep the number of computations and the number of forwards as low as possible. A straightforward way to do it is to lower the number of designs. 

\begin{figure}[h]
    \centering
    \includegraphics[width=0.8\textwidth]{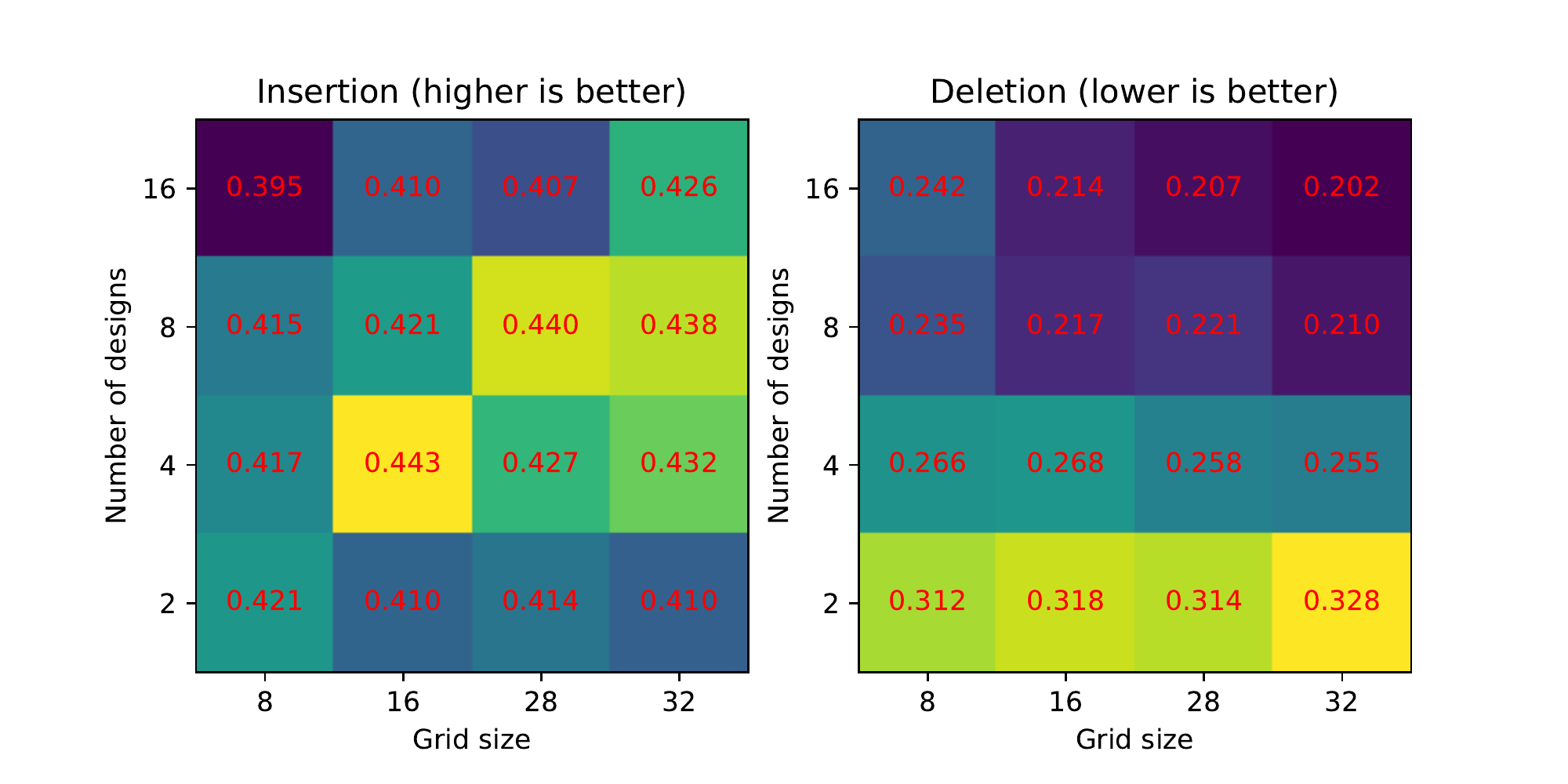}
    \caption{Insertion and deletion scores on the WCAM when the {\tt grid\_size} and {\tt nb\_design} parameters vary. For insertion, the higher, the better. For deletion, the lower, the better. Experiment carried out with a ResNet-50 \cite{he_deep_2016} backbone on 100 randomly sampled images from ImageNet.}
    \label{fig:parameters}
\end{figure}

\paragraph{Grid size and number of designs} \autoref{fig:parameters} depicts the insertion and deletion scores as a function of the {\tt grid\_size} and {\tt nb\_design}. We can notice that the highest values for insertion are obtained for a grid size of 16. It is explained by the fact that we evaluate these scores on the spatial WCAM. It evaluates the quality of the "standard" explanation without taking into account the spectral dimension of the explanations. Experiments took between a few minutes for the easiest combinations and up to 4 hours on a Google Colaboratory A100 GPU for the most computationally intensive combination.

\begin{figure}[h]
    \centering
    \includegraphics[width=\textwidth]{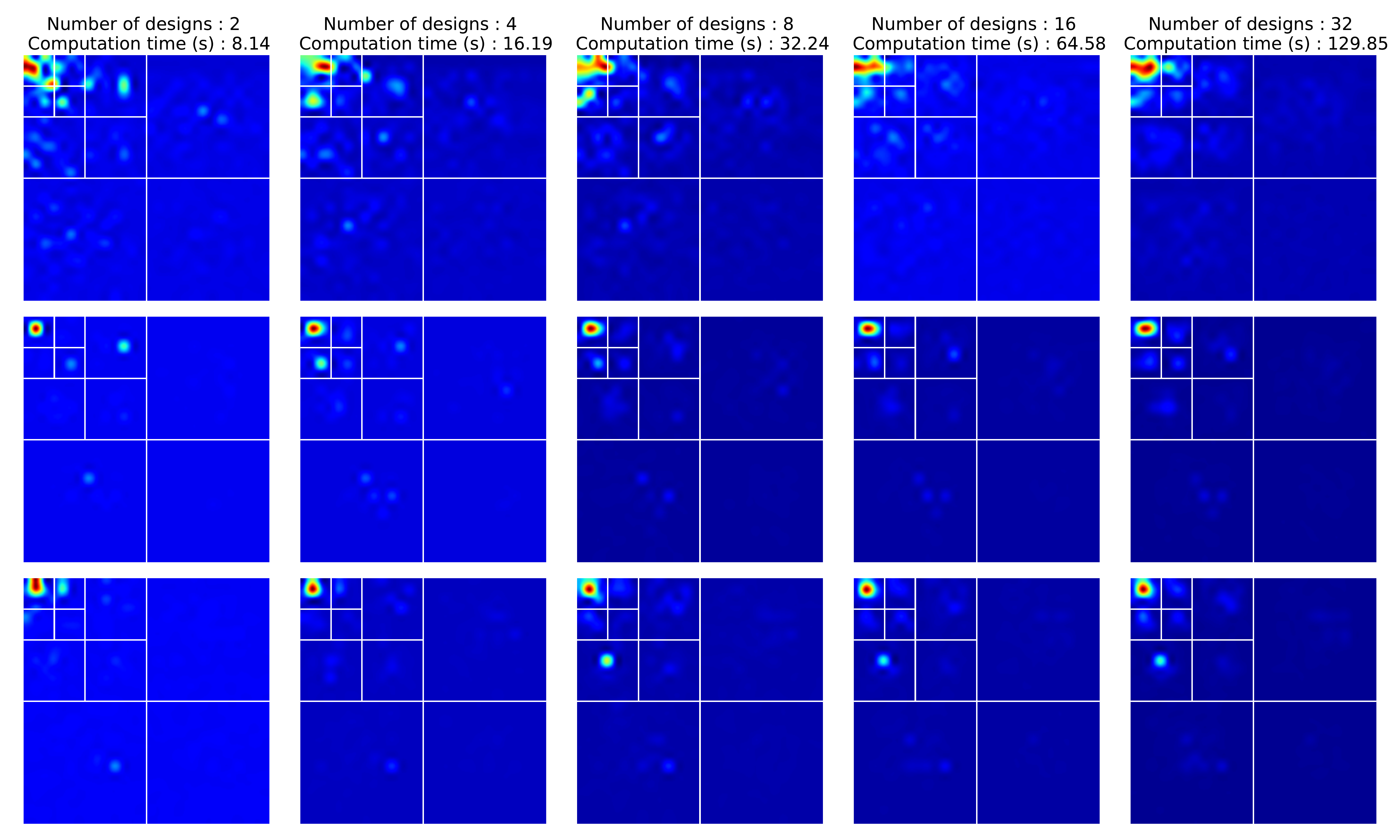}
    \caption{Effect of the number of designs on estimating the total Sobol indices. The higher the number of designs, the better the estimation of the conditional variance of the Sobol indices, but also the higher the computational cost as the number of required inferences increases. We can see that setting the number of designs to 8 is a reasonable choice to balance between accuracy and computational time. Each row depicts a different image. Each column plots the WCAM with a given number of designs (2, 4, 8, 16, and 32).}
    \label{fig:nb_design}
\end{figure}

\paragraph{Visual inspection} To further calibrate the {\tt nb\_design}, we estimated the WCAM using an increasing number of designs (columns) for different images (rows). We also report the computation time in seconds. \autoref{fig:nb_design} plots the results. We can see from this figure and \autoref{fig:parameters} that setting the number of designs to 8 is sufficient for consistently estimating the Sobol indices. This number balances between accuracy and computation time. 

\paragraph{Samplers} \autoref{tab:samplers} reports the insertion and deletion scores when the sampler changes. The baseline ScipySobolSequence gives the best results for insertion, and the Halton sequence performs slightly better for deletion. All scores are not significantly different from each other. This shows that the sampler does not affect the results.

\begin{table}[h]
    \centering
    \begin{tabular}{c c c}
    \toprule
     Sampler    &  Insertion ($\uparrow$) & Deletion ($\downarrow$) \\
     \midrule
     ScipySobolSequence \cite{sobol_distribution_1967}   & {\bf 0.440} & 0.221\\
     Halton \cite{halton_efficiency_1960} &0.438 & {\bf 0.211} \\
     Latin Hypercube \cite{mckay_comparison_1979} &0.423 & 0.217 \\
     MonteCarlo &0.435 & 0.223 \\
     \bottomrule
    \end{tabular}
    \caption{Effect of the sampler on estimating the Sobol coefficients. The {\tt grid\_size} is set to 28 and the number of designs to 8. The model backbone is a ResNet-50 \cite{he_deep_2016}.}
    \label{tab:samplers}
\end{table}

\section{Additional results}

\subsection{Additional benchmarking results}\label{sec:wcam-attribution}

\paragraph{Spatial WCAM}\label{sec:spatial-cam}

\begin{figure}[h]
    \centering
    \includegraphics[width = 0.5\textwidth]{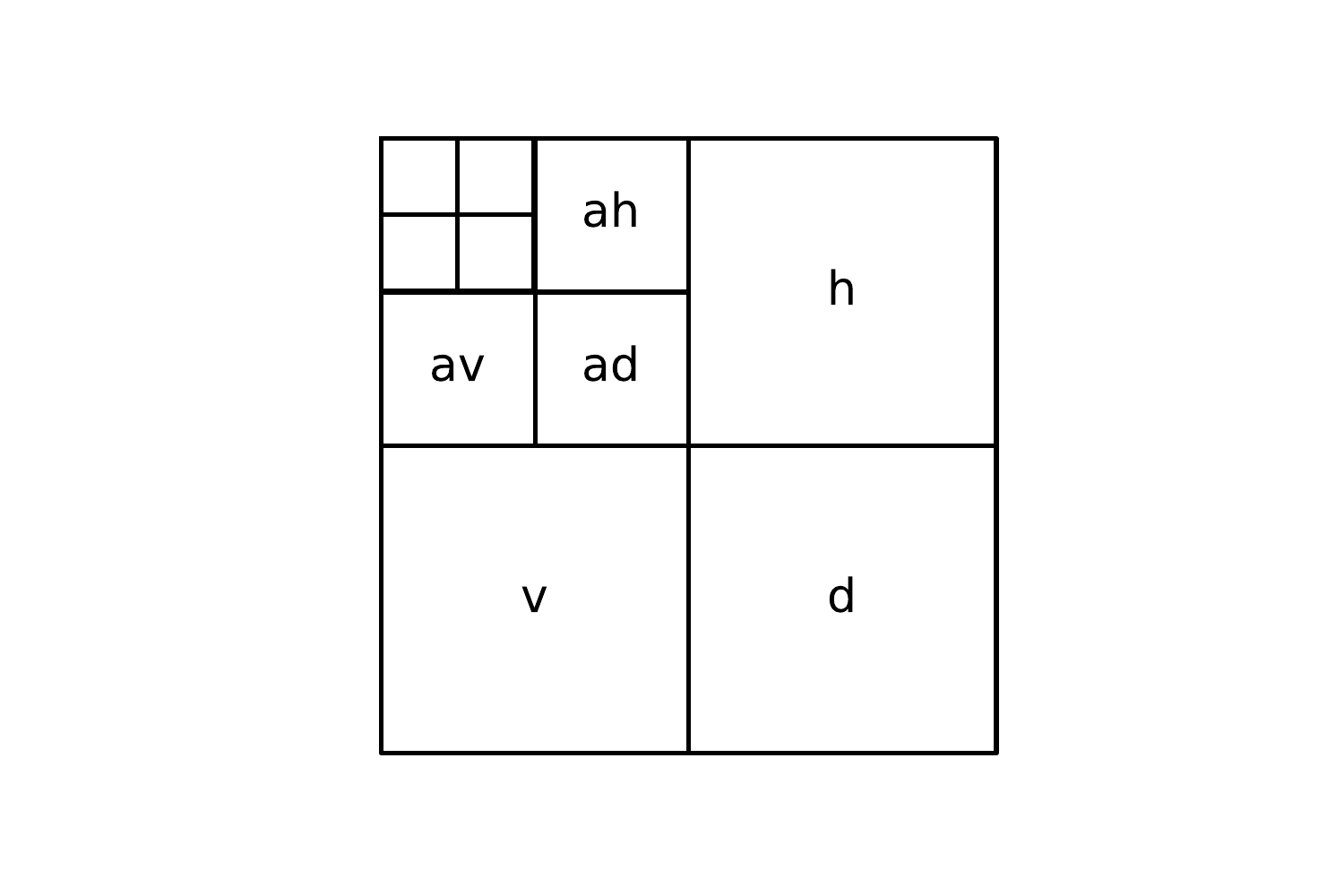}
    \caption{Decomposition of the regions of a three-level dyadic wavelet transform}
    \label{fig:decomposition-app}
\end{figure}

To recover the spatial WCAM, we sum the Sobol indices at different scales. Each subset of the wavelet transform ("h," "v," "d," "ah," "av," "ad") on \autoref{fig:decomposition-app} is indexed in space. Therefore, a point in the center of the "ad" square has the same spatial localization as a point in the center of the "h" or "v" square. 

The accuracy of the spatial cam rests on the definition of the Sobol coefficients devoted to estimating the importance of the approximation coefficients. With a $28\times28$ grid, 49 coefficients (grid size of 7) describe this level, marginally less than the default resolution of the baseline Sobol attribution method (which has a grid size of 8).

\paragraph{Insertion and deletion scores} \autoref{tab:deletion-and-insertion} reports the insertion and deletion \cite{petsiuk_rise_2018} scores compared to other attribution methods. We considered two variants of our method: the original method in the wavelet domain and the spatial WCAM, which reprojects the explanation in the pixel domain only. We can see that our method achieves state-of-the-art performance on the deletion metric and is competitive for insertion, while the spatial WCAM, which is less informative as it discards the scale information, is in line with existing methods. Benchmarks were carried out using the Xplique toolbox \cite{fel_xplique_2022} on validation images from ImageNet\cite{russakovsky_imagenet_2015} and took a couple of hours per model to compute the WCAM explanations. The computation of the baselines took another couple of hours for all models. 


\begin{table}[h]
\small
  \centering
  \vspace{0mm}\caption{\textbf{Deletion} and \textbf{Insertion} scores obtained on 100 ImageNet validation set images. For Deletion, lower is better, and higher is better for Insertion. The best results are \textbf{bolded} and second best \underline{underlined}. All benchmarks use the Xplique library \cite{fel_xplique_2022}.
  }\label{tab:deletion-and-insertion}
  \resizebox{\textwidth}{!}{\begin{tabular}{c lcccc}
  \toprule
   
   & Method & \textit{VGG16} \cite{simonyan_very_2015} & \textit{ResNet50} \cite{he_deep_2016} & \textit{MobileNet} \cite{howard_mobilenets_2017} & \textit{EfficientNet} \cite{tan_efficientnet_2020} \\
   \midrule
   Deletion ($\downarrow$)  & & & & & \\
  
  \multirow{5}{*}{White-box}
  & Saliency~\cite{simonyan_deep_2014} & 0.100  & 0.124  & 0.096  & 0.096    \\ 
  & Grad.-Input~\cite{shrikumar_not_2017} &  \underline{0.050}  & \underline{0.083}  & \underline{0.053}  & \underline{0.076}   \\
  & Integ.-Grad.~\cite{sundararajan_axiomatic_2017} & {\bf 0.041}  & {\bf 0.071}  & {\bf 0.045}  & {\bf 0.069}  \\
  & GradCAM++~\cite{selvaraju_grad-cam_2020} & 0.110  & 0.183  & 0.091  & 0.154   \\ 
  & VarGrad~\cite{selvaraju_grad-cam_2020} & 0.148  & 0.176  & 0.087  & 0.147  \\
  
  \midrule

  \multirow{3}{*}{Black-box}
  & RISE~\cite{petsiuk_rise_2018} & {\bf 0.105}  & \underline{0.143}  & {\bf 0.093}  & 0.114  \\ 
  & Sobol \cite{fel_look_2021} & 0.110  & 0.144  & \underline{0.097}  & \underline{0.101}    \\ 
  & Spatial WCAM (ours)  &0.178  & 0.221  & 0.173  & 0.185\\ 
  & WCAM (ours)  & \underline{0.107}  & {\bf 0.017} & 0.098   & {\bf 0.020} \\ 
  \toprule
   Insertion ($\uparrow$)  & & & & & \\

  \multirow{5}{*}{White-box}
    & Saliency~\cite{simonyan_deep_2014} & 0.219  & 0.232  & \underline{0.188}  & 0.164  \\ 
  & Grad.-Input~\cite{shrikumar_not_2017} & 0.140  & 0.134  & 0.119  & 0.082  \\
  & Integ.-Grad.~\cite{sundararajan_axiomatic_2017} & 0.171  & 0.170  & 0.186  & 0.147 \\
  & GradCAM++~\cite{selvaraju_grad-cam_2020} &  {\bf 0.399}  & {\bf 0.448}  & 0.084  & {\bf 0.257}   \\ 
  & VarGrad~\cite{selvaraju_grad-cam_2020} & \underline{0.223}  & \underline{0.257}  & {\bf 0.371}  & \underline{0.178}  \\

  \midrule  
  \multirow{3}{*}{Black-box}
  & RISE~\cite{petsiuk_rise_2018} & {\bf 0.460}  & {\bf 0.517}  & {\bf 0.457}  & \underline{0.402}  \\ 
  & Sobol \cite{fel_look_2021} & \underline{0.377}  & 0.428  & \underline{0.351}  & 0.294  \\ 
  & Spatial WCAM (ours)  &0.331  &0.440  & 0.326  & 0.316\\ 
    & WCAM (ours)  & 0.150 & \underline{0.460} & 0.161  & {\bf 0.737}\\ 
  \bottomrule
  \end{tabular}}
  \end{table}



\subsection{Highlighting the scale inconsistencies}\label{sec:scale-inconsistency}

\begin{definition}[Scale embedding]
    A scale embedding is a vector $z = (z_1,\dots, z_S)\in\mathbb{R}^S$ where each component $z_s$ encodes the importance of the $s^{\textrm{th}}$ scale component in the prediction. 
\end{definition}

We compute the scale embeddings by summing the Sobol indices located in each scale component. By definition, $S = 1 + 3 \times \textrm{levels}$ as we decompose each level into vertical, horizontal, and diagonal components, and we also consider the approximation coefficients. 

%

The scale embeddings represent the model's prediction in the scale domain $\mathbb{R}^S$. We measure their similarity in this space by computing the distance between two embeddings. If the distance between two embeddings is 0, the model relies on the same scales to predict the class label. By {\it relying on the same scale}, we mean that the importance of each scale was the same in both embeddings, i.e., $z_1^{(i)} = z_1^{(j)}, \dots, z_S^{(i)} = z_S^{(j)}$ for two distinct images $x_i$ and $x_j$.

On the other hand, the larger the distance, the higher the discrepancy and thus the {\it inconsistency} between predictions. Formally, we can compute the distance $d(z_i, z_j)\equiv d_{i,j}$ between two scales embeddings $z_i$ and $z_j$ corresponding to two images $i$ and $j$ of a batch $B$ of images. The average distance $\bar{d}$ informs us about the scale consistency of the batch $B$. To measure the scale consistency, we evaluate a batch $B$ of $n$ images randomly sampled from ImageNet and class-specific batches $B_1,\dots, B_c$ containing only images of the same class. We compute the average distance to evaluate the scale consistency. 

\begin{figure}[h]
    \centering
    \includegraphics[width = 0.6\textwidth]{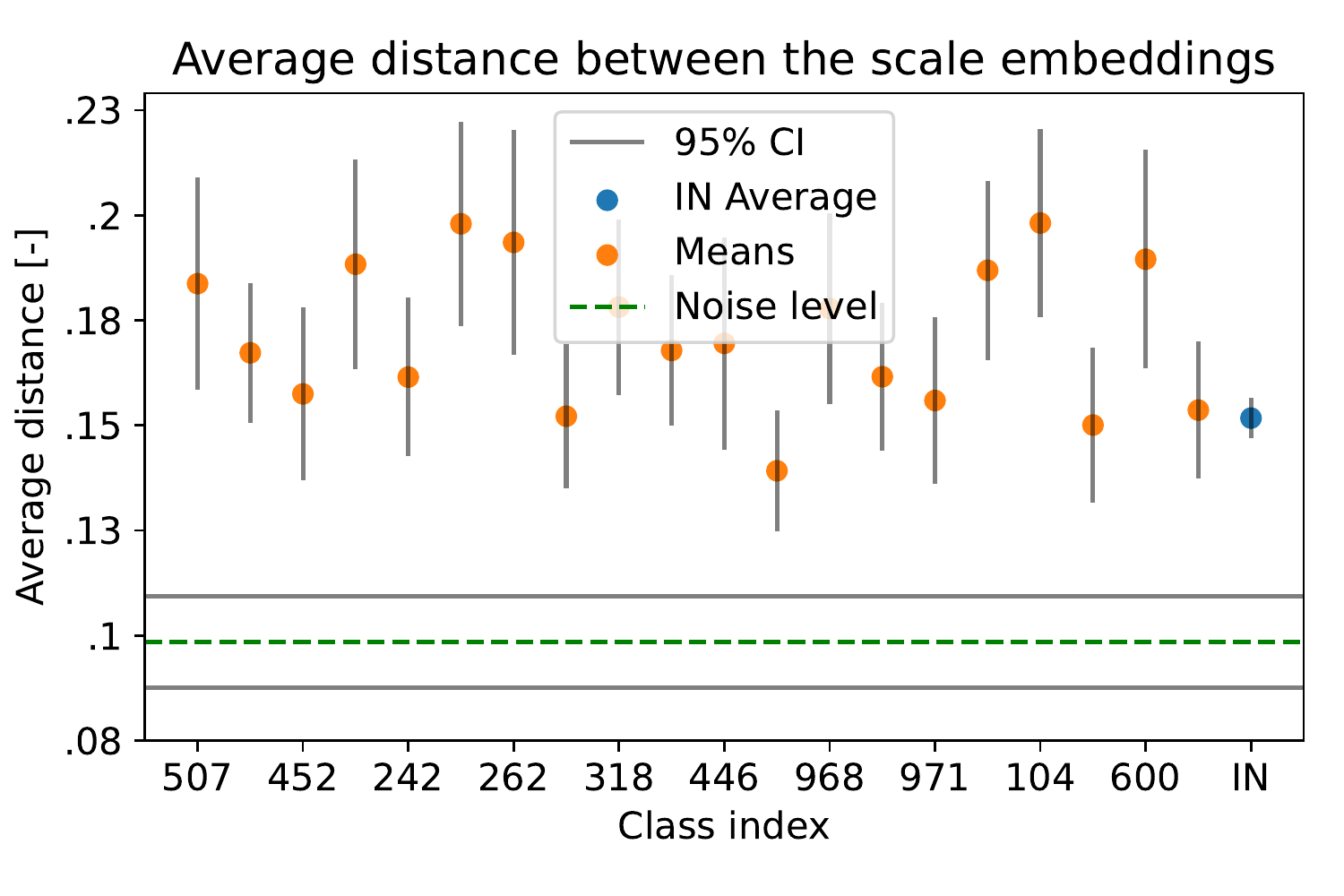}
    \caption{Average distance in the scale domain between samples from the same class. The noise line plots the background noise due to the sampling process. Continuous grey lines plot the 95\% confidence interval around the average noise caused by the sampling process.}
    \label{fig:edm_in}
\end{figure}

\paragraph{Unveiling the scale inconsistency} \autoref{fig:edm_in} plots the average Euclidean distance between two scale embeddings across 20 randomly sampled ImageNet classes. For each class, we compute the average distance between 50 scale embeddings. We can see that (i) the average difference among the classes significantly differs from the noise attributable to randomness (bottom green dotted line). (ii) We can see that the distance between two random ImageNet (rightmost dot) images is not different from intra-class distances. We interpret this result as evidence that when shifting from one instance to the other, the model does not necessarily rely on the same scale, i.e., on the same factors to make its prediction. 

\paragraph{Mechanisms} The scale inconsistency we uncover in the wavelet domain is reminiscent of the {\it frequency shortcuts} highlighted in a concurrent work \cite{wang_what_2023}. This work showed that learning shortcuts \cite{geirhos_shortcut_2020} can be characterized in the frequency domain: models tend to favor the most distinctive frequencies to make a prediction. Crucially, \cite{wang_what_2023,wang_frequency_2022} also document that these shortcuts are context-dependent. Our results provide a more general characterization of this phenomenon, as it affects not only CNNs but also transformers, and data augmentation techniques cannot mitigate it. Therefore, the reason behind these frequency shortcuts appears to be the optimization process. In that light, recent work \cite{pezeshki_gradient_2021} introduced the notion of {\it gradient starvation} phenomenon to explain learning shortcuts. According to this theory, "{\it when a feature is learned faster than the others, the gradient contribution of examples that contain this feature is diminished}." \cite{puli_dont_2023} also argued that the cause for shortcut learning lies in the gradients and cross-entropy.

\subsection{Reconstruction based on the most important Wavelet coefficients}\label{sec:reconstruction}

\begin{figure}[h]
    \centering
    \includegraphics[width = .9\textwidth]{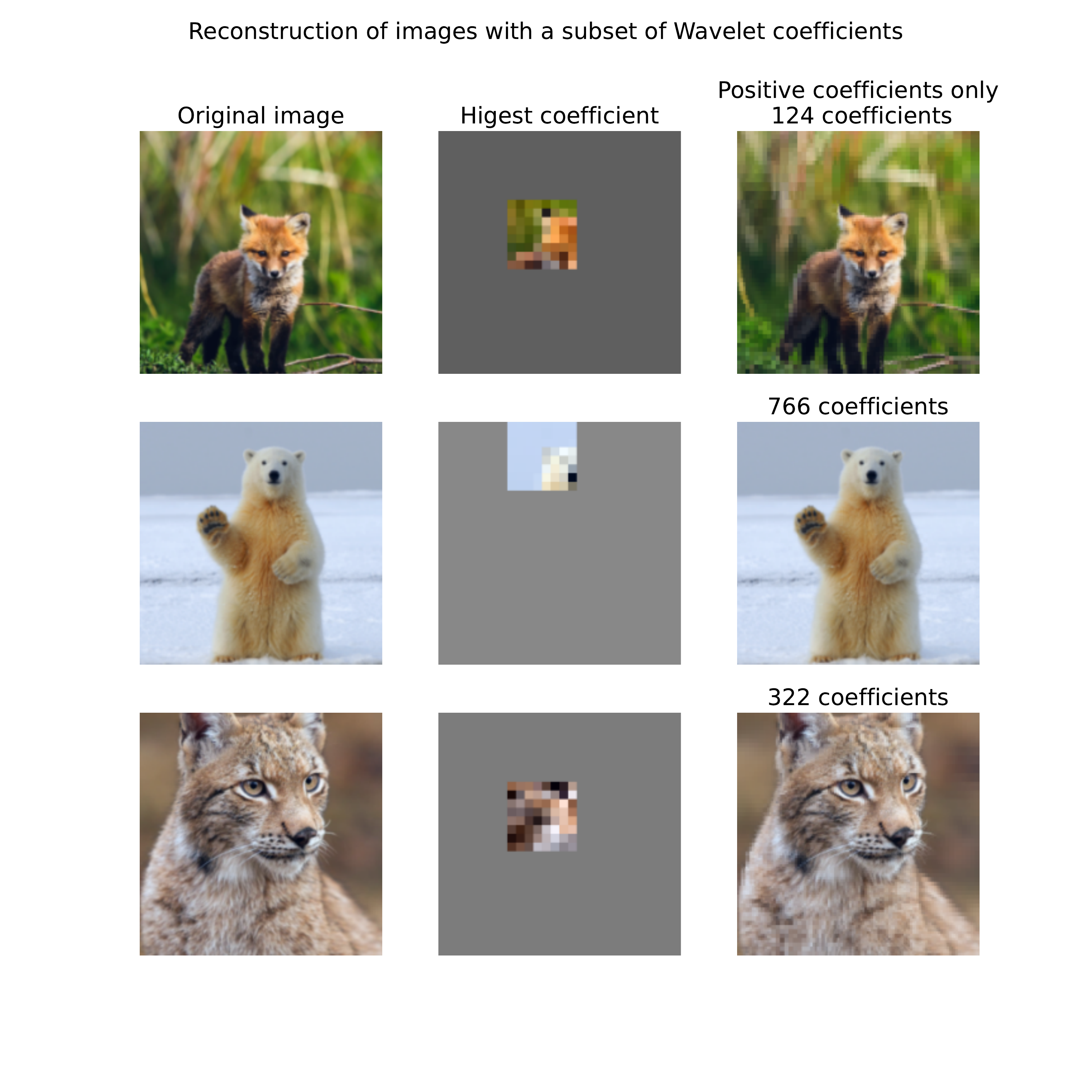}
        \caption{Example of reconstruction using the most important Wavelet coefficients. On the column in the center, we reconstruct the image with only the most important coefficient. On the rightmost column, we plot the image reconstructed with only the $n$ positive coefficients.}
    \label{fig:reconstruction-example}
\end{figure}

The WCAM points towards the most important wavelet coefficients. We leverage the WCAM to see how many coefficients we need to reconstruct an image that the model can correctly classify. \autoref{fig:reconstruction-example} illustrates the reconstruction from only the most important wavelet coefficients. Wavelet coefficients are ranked using their associated Sobol indices. We reconstruct images with a gradually increasing number of coefficients. 

\begin{figure}[h]
    \centering
    \includegraphics[width=.5\textwidth]{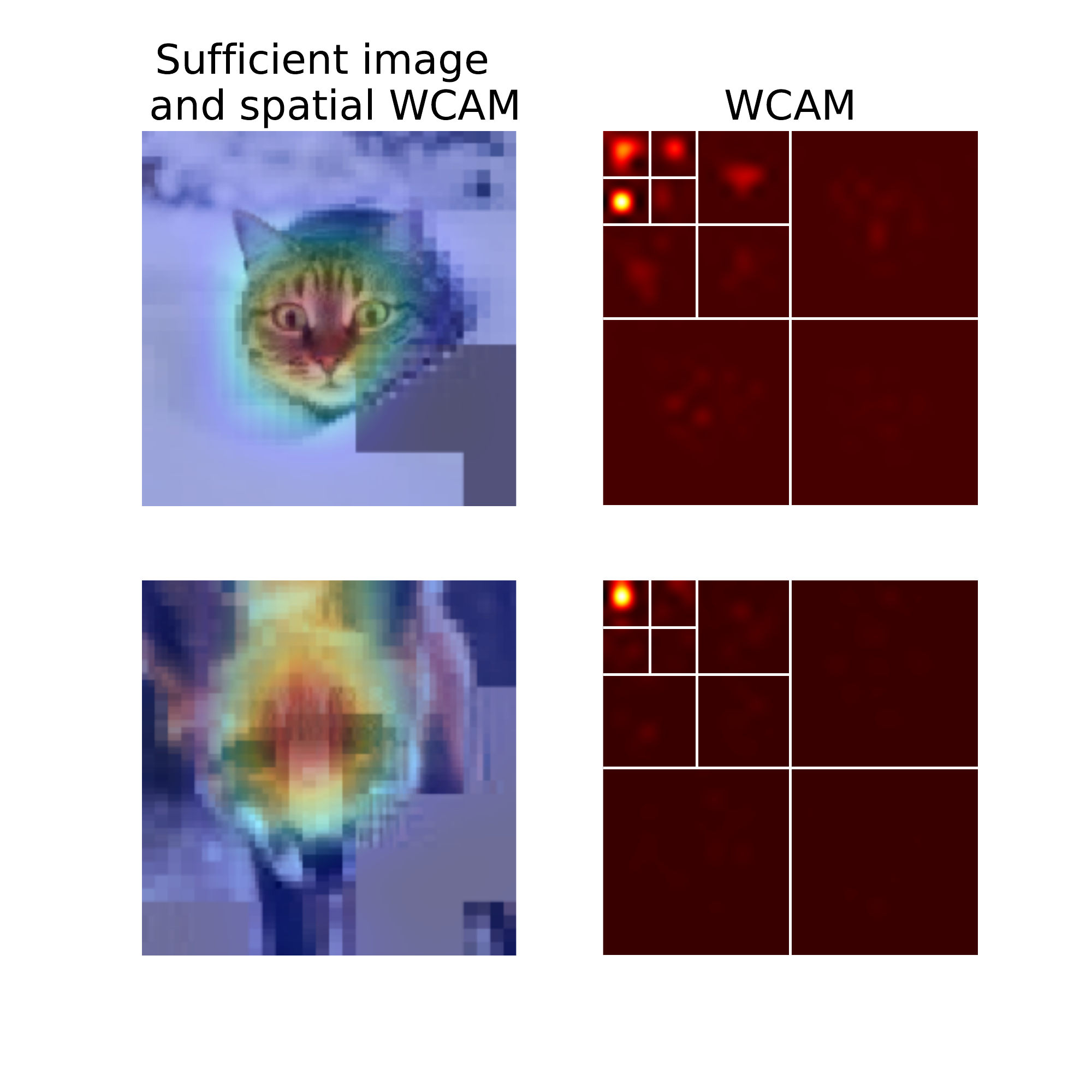} 
\caption{Sufficient or minimal images reconstructed from the WCAM.}
\label{fig:reconstruction-important-coeffs}
\end{figure}

\paragraph{Minimal image} We call {\it minimal} (or sufficient) image the image reconstructed from the $n$ first wavelet coefficients according to their corresponding Sobol indices such that the model can correctly predict the image's label. \autoref{fig:reconstruction-important-coeffs} displays examples of such images. In our examples, we can see that for the image of a cat, the model needs detailed information around the eyes, which is not the case for the fox. In both cases, we also see that the models do not need information from the background, as we can completely hide it without changing the prediction. Identifying a minimal image could have numerous applications, for instance, in transfer compressed sensing \cite{dhar_modeling_2018}. In some applications (e.g., medical imaging), data acquisition is expensive \cite{dhar_modeling_2018}. Using the information provided by the WCAM, we can learn how to compress signals to preserve the essential information for classification. 

\newpage

\autoref{fig:reconstruction-important-coeffs} shows that the minimal image varies from one sample to the other. In some cases, only coarse details are necessary, while in other cases, fine-grained details are necessary for the prediction. The minimal image highlights what the model needs to see to predict a label correctly. Assessing whether this image contains the relevant factors could be helpful for the practitioner to check if the model abides by the expectations.

\end{document}